\pdfoutput=1

\documentclass[11pt]{article}

\usepackage[]{latex/acl}

\usepackage{times}
\usepackage{latexsym}

\usepackage[T1]{fontenc}

\usepackage[utf8]{inputenc}

\usepackage{microtype}

\usepackage{inconsolata}

\usepackage{multirow}
\usepackage{booktabs}
\usepackage{arydshln}
\usepackage{caption}
\usepackage{graphicx}
\usepackage{amsmath}
\usepackage{amssymb}
\usepackage{colortbl}
\usepackage{xcolor} 
\definecolor{highlightcolor}{rgb}{0.9, 0.9, 0.9} 
\title{What Have We Achieved on Non-autoregressive Translation?}

\author{ 
 Yafu Li$^{\spadesuit \clubsuit}\footnotemark[1]$\hspace{0.5mm},
 Huajian Zhang$^{\clubsuit}\footnotemark[1]$\hspace{0.5mm},
 Jianhao Yan$^{\clubsuit}$\hspace{0.5mm} \\
  \bf{
   Yongjing Yin$^{\clubsuit}$\hspace{0.5mm},
 Yue Zhang$^{\clubsuit }$\footnotemark[2]}\hspace{0.2mm}\hspace{1.5mm} \\
$^\spadesuit$ Zhejiang University \ \ \ \quad$^\clubsuit$Westlake University \\
 \texttt{\{yafuly,huajian.zhang98\}@gmail.com}  \\
 \quad\texttt{\{yanjianhao,yinyongjing,zhangyue\}@westlake.edu.cn}\\
}

\begin{document}
\maketitle
\renewcommand{\thefootnote}{\fnsymbol{footnote}}
\footnotetext[1]{\ Equal contribution.}
\footnotetext[2]{\ Corresponding author.}

\begin{abstract}
Recent advances have made non-autoregressive (NAT) translation comparable to autoregressive methods (AT). However, their evaluation using BLEU has been shown to weakly correlate with human annotations.
Limited research compares non-autoregressive translation and autoregressive translation comprehensively, leaving uncertainty about the true proximity of NAT to AT.
To address this gap, we systematically evaluate four representative NAT methods across various dimensions, including human evaluation.
Our empirical results demonstrate that despite narrowing the performance gap, state-of-the-art NAT still underperforms AT under more reliable evaluation metrics.
Furthermore, we discover that explicitly modeling dependencies is crucial for generating natural language and generalizing to out-of-distribution sequences.

\end{abstract}

\section{Introduction}





Non-autoregressive translation, where the model generates translations in parallel, demonstrates notable decoding speed advantages compared with traditional autoregressive translation~\cite{VaswaniSPUJGKP17} and large language models for translation~\cite{gpt4}.
However, it suffers from performance degradation compared to autoregressive counterparts~\cite{nat}.
The degradation stems from the independence assumption, which ignores the inter-token language dependency on the target side.
Various methods are proposed to mitigate the performance gap~\cite{maskp,glat,ctc,oaxe,mgmo,da,cmlmc}.

Although representative methods~\cite{ctc,mgmo,da,cmlmc} have reported comparable translation performance to AT, almost all NAT methods are evaluated under BELU scores~\cite{bleu}.
Although BLEU has been long adopted, recent work~\cite{stop-bleu} argues that it is not a reasonable metric, considerably underperforming 
alternative metrics such as COMET~\cite{comet} or large language model evaluation~\cite{gpt4-eval}.
Limited work has been devoted to a systematic evaluation of advanced NAT against AT, leaving a significant gap in the research literature.

To address this gap, we conduct a comprehensive evaluation of representative NAT methods, aiming to reveal existing limitations and provide insights for future research.
Our primary focus is on fully non-autoregressive methods which generate translations in a one-shot manner, achieving the most decoding efficiency advantage.
We consider MgMO~\cite{mgmo} for advanced optimization, CTC~\cite{ctc} for modeling latent alignment, and DAT~\cite{da} for explicit target-side dependency modeling. 
CMLM~\cite{maskp} is adopted as the representative iterative NAT method.
All models are tested on representative benchmark datasets under a comprehensive evaluation
, including rule-based metrics, model-based metrics and GPT4-based metrics~\cite{gpt4-eval}.
Moreover, we conduct human evaluation under the MQM framework~\cite{mqm} to gain further insights into the performance of NAT models that may be overshadowed by global automatic evaluations.

Automatic evaluation demonstrates varying degrees of advantage for AT over NAT models.
In general, DAT achieves the most competitive performance, followed by MgMO and CTC.
Under rule-based evaluation metrics such as BELU and chrf~\cite{chrf},
DAT can achieve comparable or even superior performance compared to AT.
However, this competitiveness diminishes when using model-based metrics such as COMET~\cite{comet} or GPT4-based evaluation, under which AT significantly outperforms all NAT models.
Fine-grained human evaluation indicates that NAT models incorporating explicit dependency modeling (e.g., DAT and CMLM) achieve similar levels of translation fluency with AT, yet suffering various translation accuracy errors. 
Compared with AT, NAT tends to produce more \textit{grammar} or \textit{punctuation} errors.
Models without explicit dependency modeling (MgMO and CTC) suffer the most \textit{mistranslation} and \textit{omission} errors.
On the other hand, models with latent alignments (CTC and DAT) are more prone to \textit{spelling} and \textit{addition} errors.


Most of these errors are due to NAT's inadequate dependency modeling.
Specifically, DAT's addition errors occur when it generates repeated translations, known as n-gram repetition.
This can be easily overlooked by BLEU evaluation, which measures n-gram precision, explaining why DAT performs well in terms of BLEU but not COMET.
The n-gram repetition mainly stems from the weak, though explicit, dependency modeling.
DAT limits inter-token dependency within one step using a one-linear-layer attention module for decoding efficiency.
In contrast, AT can depend on the entire generation history and encode it with powerful Transformer blocks.
To validate our assumption, we train an asymmetric AT with a one-layer decoder and observe similar n-gram repetitions.
Furthermore, adding an additional linear layer to the transition attention in DAT effectively reduces the repetition, corroborating our hypothesis.

Apart from translation quality, we compare AT with NAT from the perspective of generalization and robustness. Empirical findings demonstrate that explicit dependency modeling is crucial for generating human-like languages and generalizing to out-of-distribution samples, which NAT methods lack or are still weak at.
On the other hand, weak dependency exhibits stronger robustness to input perturbations, as it is less affected by exposure bias~\cite{exposure1,exposure2}.
%
Future research on NAT should focus on how to consolidate explicit language dependency while maintaining decoding efficiency.
We release our resources at \url{https://github.com/HJZnlp/NAT_vs_AT}.

\section{Method}
\label{sec:method}
We begin with a brief introduction to autoregressive and non-autoregressive machine translation, before introducing four representative NAT methods.

\subsection{Neural Machine Translation}
The machine translation task can be formally defined as a sequence-to-sequence generation problem, where the model generates the target language sequence $\mathbf{y} = \{y_1,y_2,...,y_T\}$ from the target vocabulary $\mathcal{V}$, given the source language sequence ${\mathbf{x} = \{x_1,x_2,...,x_S\}}$ based on the conditional probability $p_{\mathbf {\theta}}(\mathbf{y}|\mathbf{x})$ ($\mathbf {\theta}$ denotes the model parameters).
\paragraph{Autoregressive Translation.}
Autoregressive neural machine translation factorizes the conditional probability to $\prod_{i=1}^{T}p(y_i|y_1,...,y_{t-1},\mathbf{x})$, where the model is trained in a teacher-forcing way with cross-entropy (XE):
\begin{equation}
    \mathcal{L}_{\text{AT}} = - \log p(\mathbf{y}|\mathbf{x}) = -\sum_{i=1}^{T}\log p_{\mathbf{\theta}}(y_i|\mathbf{x},y_{<i}).
\end{equation}
During inference, the model sequentially generates tokens based on previous predictions.

\paragraph{Non-autoregressive Translation.} In contrast, non-autoregressive machine translation \cite{nat} ignores the dependency between target tokens and factorizes the probability as $\prod_{i=1}^{T}p(y_i|\mathbf{x})$, where tokens at each time step are predicted independently. 
Vanilla NAT models are optimized with XE loss with target dependency ignored:
\begin{equation}
\label{eq:nat}
    \mathcal{L}_{\text{NAT}} = - \log p(\mathbf{y}|\mathbf{x}) = -\sum_{i=1}^{T}\log p_{\mathbf{\theta}}(y_i|\mathbf{x}),
\end{equation}
with an additional loss for length prediction:
\begin{equation}
    \mathcal{L}_{\text{length}} = -\log p_{\mathbf{\theta}}(T|\mathbf{x}).
\end{equation}

\paragraph{Challenges of NAT.} 
The major difficulty of non-autoregressive translation lies in that the decoder side relies solely on the source-side information without any target inputs, e.g., history predictions in AT.
Autoregressive models utilize previous token predictions to select the next token from the distribution over the whole vocabulary space:
\begin{equation}
\label{eq:at}
    p_{\mathbf{\theta}}(y_i|y_{<i},\mathbf{x}) = \text{softmax}(\mathbf{W}_\text{P}\text{Transformer}(y_{<i},\mathbf{x}),
\end{equation}
where $\mathbf{W}_\text{P}$ is the vocabulary projection weight.
The inter-token dependency involves layers of Transformer blocks.
In contrast, NAT models generate translations in a "one-shot" manner, ignoring or weakening the strong language dependency on the target side.
As a result, vanilla NAT is not capable of properly modeling the highly multi-modal distribution of target translations, i.e., a source sentence can have multiple valid translations.
Various methods aim to alleviate the conditional independence assumption.
In this work, we consider four representative methods: (1) alternative optimization with model architecture unchanged~\cite{mgmo}; (2) introducing latent alignments based on an upsampled decoder prediction~\cite{ctc}; (3) building shallow but explicit target-side dependency~\cite{da}; and (4) iterative decoding~\cite{maskp}.

\subsection{NAT with Advanced Optimization}
Instead of exerting token-by-token cross-entropy supervision, ~\citet{mgmo} propose multi-granularity optimization (MgMO) to collect multi-granularity feedback on generations sampled from the models and gather them for backpropagation:
\begin{equation}
\label{eq_loss_mgmo}
    \mathcal{L}_{\text{MO}} =-\sum_{k=1}^{K}q_{\mathbf {\theta}}(\mathbf{h}^k|\mathbf{x})R(\mathbf{h}^k,\mathbf{y}^k),
\end{equation}
where $K$ is the sample space size.
$q_{\mathbf {\theta}}(\mathbf{h}^k|\mathbf{x})$ is defined as the normalized probability for each hypothesis $\mathbf{h}^k$:
\begin{equation}
    q_{\mathbf{\theta}}(\mathbf{h}^k|\mathbf{x};\alpha) = \frac{\hat{p}_{\mathbf {\theta}}(\mathbf{h}^k|\mathbf{x})^{\alpha}}{\sum_{\mathbf{h^{\prime}}\in \mathcal{K}(\mathbf{x})}\hat{p}_{\mathbf{\theta}}(\mathbf{h^{\prime}}|\mathbf{x})^\alpha},
\end{equation}
where $\mathcal{K}(\mathbf{x})$ denotes the sample space and $\alpha$ controls the distribution sharpness.
$R(\mathbf{h},\mathbf{y})$ is a reward function that encourages the generations to be similar with references under various granularity.
MgMO requires no architecture modification and thus maintains decoding efficiency.

\subsection{NAT with Latent Alignments}

\citet{ctc} introduce latent alignment models, e.g., Connectionist Temporal Classification (CTC)~\cite{speech}, to mitigate the target-side independence assumption.
CTC utilizes a sequence of discrete latent alignment variables to monotonically align the non-autoregressive predictions of the model and target side tokens. 
The marginal probability over latent alignments $\mathbf{a}$ is derived as:
\begin{equation}
 \begin{aligned}
 \mathcal{L}_{\text{LA}} &= -\log p_{\mathbf{\theta}}(\mathbf{y}|\mathbf{x}) \\
  &= -\log \sum_{\mathbf{a} \in \beta(\mathbf{y})} p_{\mathbf{\theta}}(\mathbf{y}|\mathbf{a},\mathbf{x})p_{\mathbf{\theta}}(\mathbf{a}|\mathbf{x}),
 \end{aligned}
\end{equation}
where $\beta(\mathbf{y})$ is a function that returns all possible alignments for a sequence $\mathbf{y}$.
Then $\mathbf{a}=\{a_1,\ldots,a_{M}\}$ is predicted by the decoder output states $\mathbf{H}=\{\mathbf{h}_1,\ldots,\mathbf{h}_M\}$, 
where $a_i\in \mathcal{V} \cup \{\text{``}\_\text{''}\}$.
``\_'' is a special blank token to allow many-to-one and null alignment.
For instance, for a target sequence ``thank you'', valid alignments $\mathbf{a}$ include ``\_ thank thank you'' and ``thank \_ you \_''.
The decoder state length is set as several times the source sequence length to allow long translations.
The alignment probability
$p_{\mathbf{\theta}}(\mathbf{a}|\mathbf{x})$ 
is derived by:
\begin{align}
\label{eq_alignment}
    p_{\mathbf{\theta}}(\mathbf{a}|\mathbf{x})&= \prod_{i=1}^{M} p_{\mathbf{\theta}}(a_i|\mathbf{x}) \notag \\
    &= \prod_{i=1}^{M} \text{softmax}(\mathbf{W}_\text{P} \mathbf{h}_i). 
\end{align}
Since $a_i\in \mathcal{V} \cup \{\text{``}\_\text{''}\}$, the posterior probability of $\mathbf{y}$ becomes:
\begin{equation}
p_{\mathbf{\theta}}(\mathbf{y}|\mathbf{a},\mathbf{x}) = \begin{cases} 
1 & \text{if } \mathbf{a} \in \beta(\mathbf{y}) \\
0 & \text{otherwise}.
\end{cases}
\end{equation}

MgMO and CTC avoid token-by-token CE supervision by introducing segment-level optimization or marginalizing latent alignments.
However, they suffer independence assumption in generating tokens (Equation~\ref{eq:nat}) or alignments (Equation~\ref{eq_alignment}).
Consequently, both MgMO and CTC cannot inherently handle multi-modal problems and heavily rely on techniques such as knowledge distillation~\cite{understanding_kd} to mitigate this limitation.

\subsection{NAT with Explicit Dependency}
\citet{da} propose directed Acyclic Transformer (DAT) to construct explicit dependencies, by formalizing an alignment as a path in a direct acyclic graph.
Similar to CTC, the decoder state length is upsampled to $M$ and $\mathbf{H}=[\mathbf{h}_1, \ldots, \mathbf{h}_{M}]$ denotes the decoder output hidden states, which are defined as the vertex states.
The probability of path $\mathbf{a}$
is redefined as the position transition probability:
\begin{gather}
    p_{\mathbf{\theta}}(\mathbf{a}|\mathbf{x}) = \prod_{i} p_{\mathbf{\theta}}(a_{i+1}|a_{i}, \mathbf{x}) =  \prod_{i} \mathbf{\mathbf{E}}_{a_{i}, a_{i+1}}, \notag
\end{gather}
where $\mathbf{E} \in \mathbb{R}^{M \times M}$ is the transition matrix normalized by rows. 
$\mathbf{a}=\{ a_1, a_2, \ldots, a_T \}$ is a possible path represented by a sequence of vertex indexes of the vertex states $\mathbf{H}$, i.e., $a_i \in \{1, 2, 3, \ldots, M\}$.
Specifically, the transition matrix is obtained by:
\begin{gather}
    \mathbf{E} = \text{softmax}(\frac{\mathbf{Q} \mathbf{K}^T}{\sqrt{d}}), \label{eq_qk} \\
    \mathbf{Q} = \mathbf{H}  \mathbf{W}_\text{Q},\quad \mathbf{K} = \mathbf{H} \mathbf{W}_\text{K},  \notag
\end{gather}
where $d$ is the hidden size, $\mathbf{W}_\text{Q}$ and $\mathbf{W}_\text{K}$ are learnable matrices.
Conditioned on the vertex states in $\mathbf{H}$ and the selected path $\mathbf{a}$, the posterior probability of $\mathbf{y}$ is computed as:

\begin{align}
p_{\mathbf{\theta}}(\mathbf{y}|\mathbf{a}, \mathbf{x}) &= \prod_{i=1}^{T} P_{\mathbf{\theta}}(y_i|a_i, \mathbf{x}) \notag\\
&= \prod_{i=1}^{T} \text{softmax}(\mathbf{W}_\text{p} \mathbf{h}_{a_i}),
\end{align}
where $\mathbf{h}_{a_i}$ is the representation of the $i$-th vertex on the path $\mathbf{a}$.

Different from previous NAT methods, DAT explicitly models token dependencies through vertex transitions. 
DAT first parallelly predicts a subset of all possible tokens for translating the source sentence
and stores it as $\mathbf{H}$, whose size is usually several times (e.g., 8) that of the source sequence.
In contrast to Equation~\ref{eq:at}, the inter-token dependency is a one-step local transition for each vertex $\mathbf{h_i}$, to determine the next token from the rest of the set, i.e., $\{\mathbf{h_{i+1},\ldots,h_{M}}\}$:
\begin{gather}
p_{\mathbf{\theta}}(\mathbf{y}) = \prod_{i=1}^{T} p_{\mathbf{\theta}}(y_{a_i}|y_{a_{i-1}}),  \\
   p_{\mathbf{\theta}}(y_{a_i}|y_{a_{i-1}}) = \text{softmax}(\mathbf{W}_\text{P}\mathbf{h}_{\text{argmax}(\mathbf{\mathbf{E}}_{a_{i-1}, a_{i}})}),
\end{gather}
where $y_{a_i}$ is the predicted token of the $i$-th vertex on the path $\mathbf{a}$\footnote{We omit conditional dependency on $\mathbf{x}$ for simplicity.}.
The explicit though weak dependency modelled by one-layer linear weights $\mathbf{W}_{\text{Q}}$ and $\mathbf{W}_{\text{K}}$ alleviate the necessity of knowledge distillation, yet suffering n-gram repeating issues (discussed in Section~\ref{sec:explicit_dependency}).


\subsection{NAT with Iterative Refinement}
The iterative NAT model~\cite{maskp} is typically trained with conditional masked language modeling (CMLM) to build inter-token dependencies:
\begin{equation}
\label{eq_cmlm}
    \mathcal{L}_{CMLM} = -\sum_{y_t\in \mathcal{Y}(\mathbf{y})}logp_{\mathbf{\theta}}(y_t|\Omega(\mathbf{y},\mathcal{Y}(\mathbf{y})),\mathbf{x}),
\end{equation}
where $\mathcal{Y}(\mathbf{y})$ is a randomly selected subset of target tokens and $\Omega$ denotes a function that masks a selected set of tokens in $\mathcal{Y}(\mathbf{y})$.
During decoding, starting from a sequence of initiative tokens, e.g., ``<unk>'', CMLM models iteratively refine translations from previous iterations to generate target language sequences.

\begin{table*}[h!]
\centering
\small
\renewcommand{\arraystretch}{1.0} 
\setlength{\belowcaptionskip}{-0.1cm}
\begin{tabular}{lcccccc}
\toprule
Model  & \multicolumn{1}{l}{BLEU$\uparrow$}&\multicolumn{1}{l}{chrf$\uparrow$} & \multicolumn{1}{l}{COMET$\uparrow$}&\multicolumn{1}{l}{BLEURT$\uparrow$} &\multicolumn{1}{l}{GEMBA$\uparrow$} &\multicolumn{1}{l}{Speed$\uparrow$}\\
\midrule
\multicolumn{7}{c}{\textbf{WMT16 En$\Rightarrow$Ro}} \\
\midrule
\multicolumn{7}{c}{w/o Knowledge Distillation} \\
\midrule
AT~\cite{VaswaniSPUJGKP17} & \underline{\textbf{34.39}$\dagger$} & \underline{\textbf{58.48}} & \underline{\textbf{78.90}$\dagger$} & \underline{\textbf{69.89}$\dagger$}  & \underline{\textbf{86.18}$\dagger$ }&1.0$\times$
\\
NAT~\cite{nat}   &  23.75 &  50.72&  65.78 & 53.91 &67.29&15.9$\times$\\
MgMO~\cite{mgmo}  &  30.97 & 56.65 & 73.54 & 63.19 &80.04 &14.9$\times$\\
CTC~\cite{ctc}  &\underline{32.73}  & \underline{57.77} & \underline{74.26} &  \underline{63.99}& \underline{81.16}&14.5$\times$\\
DAT~\cite{da}  & \underline{33.18} & \underline{57.35} &\underline{76.14}  &  \underline{66.72}&\underline{83.72} &13.8$\times$\\
CMLM~\cite{maskp}  & 31.97 & 56.78 & 74.11 & 63.34 &78.72 &2.7$\times$\\
\midrule

\multicolumn{7}{c}{w/ Knowledge Distillation} \\
\midrule
AT~\cite{VaswaniSPUJGKP17}   & \underline{\textbf{33.92}}  & \underline{\textbf{58.45}} &  \underline{\textbf{78.49}$\dagger$} & \underline{\textbf{69.26}$\dagger$ }& \underline{\textbf{86.22}$\dagger$ }&1.0$\times$\\
NAT~\cite{nat}   & 30.97 & 56.52 & 72.60 & 62.12 & 77.47& 15.8$\times$\\
MgMO~\cite{mgmo} & 32.86 & 57.40 & 75.52 &65.36  &82.73 & 14.9$\times$\\ 
CTC~\cite{ctc}    &\underline{33.28}  & \underline{58.28} & \underline{75.54} & \underline{65.71} & \underline{82.94}&14.5$\times$\\
DAT~\cite{da}   &  \underline{33.25} & \underline{57.89} & \underline{76.59} & \underline{67.01} & \underline{84.27}&13.7$\times$\\
CMLM~\cite{maskp} & 32.71 & 56.76 & 72.36 & 63.42 &76.67 &2.7$\times$\\
\midrule
\multicolumn{7}{c}{\textbf{WMT21 De$\Rightarrow$En}} \\
\midrule
\multicolumn{7}{c}{w/o Knowledge Distillation} \\
\midrule
AT~\cite{VaswaniSPUJGKP17} & \underline{\textbf{31.89}} & \underline{\textbf{60.25}} &\underline{\textbf{84.26$\dagger$}}
&\underline{\textbf{71.94$\dagger$}}
&\underline{ \textbf{92.91$\dagger$}}  
&1.0$\times$ \\
NAT~\cite{nat}  & 16.85 & 43.46   &55.87&44.80&36.94  &15.5$\times$\\
MgMO~\cite{mgmo} & 28.89
&\underline{58.11} & \underline{77.76}&64.08&\underline{83.51}  &13.8$\times$\\
CTC~\cite{ctc}  & 27.35 &56.53 & 75.38& 61.79&79.14  &13.5$\times$ \\
DAT~\cite{da}  & \underline{31.69} &\underline{59.60} &\underline{81.12}&\underline{69.00}&\underline{88.29}  &13.1$\times$ \\
CMLM~\cite{maskp} & \underline{29.36} & 57.79 &76.39&\underline{65.00}&82.07  &2.4$\times$\\
\midrule
\multicolumn{7}{c}{w/ Knowledge Distillation} \\
\midrule
AT~\cite{VaswaniSPUJGKP17}  & \underline{32.04} &\underline{\textbf{60.85}}
&\underline{\textbf{84.72$\dagger$}}
& \underline{\textbf{72.53}$\dagger$ }
&\underline{\textbf{93.39$\dagger$}} 
& 1.0$\times$
\\
NAT~\cite{nat}  &27.55&56.56 &75.50&62.72&76.89 & 15.2$\times$ \\
MgMO~\cite{mgmo}  & 30.32&59.36 &\underline{81.15}&\underline{67.76} &\underline{89.17}  &13.8$\times$\\
CTC~\cite{ctc}    & \underline{30.52}&\underline{59.83}   &80.06&67.24& 86.91  &13.4$\times$\\
DAT~\cite{da}   &\underline{\textbf{32.26}}
&\underline{60.80} 
&\underline{83.32}&\underline{71.44}
& \underline{92.05}  & 13.1$\times$\\
CMLM~\cite{maskp} & 30.25& 58.40 &77.14&65.63& 82.98  &2.4$\times$\\
\bottomrule

\end{tabular}
\caption{\label{tab:main_merge}
Automatic evaluation results of different translation models on WMT16 En$\Rightarrow$Ro and WMT21 De$\Rightarrow$En, considering both raw data and distillation data settings.
We encompass a wide range of metrics including rule-based metrics (BLEU and chrf), model-based metrics (COMET and BLEURT) and LLM-based metrics (GEMBA).
Bold numbers represent the best performance and underlined numbers denote the top 3 performance.
$\dagger$ denotes translation quality of AT is significantly better than all other NAT models with a $p<0.01$ \cite{sigtest}.
}
\end{table*}

\section{Experiment and Setup}
\paragraph{Datasets and Models.}
We conduct experiments on WMT16 En$\Rightarrow$Ro and WMT21 De$\Rightarrow$En with 4 representative NAT methods apart from the vanilla NAT and AT.
For knowledge distillation,
We train an autoregressive model on the raw data as the teacher model to generate the distilled dataset.
To achieve peak performance, we employ glancing training~\cite{glat} for CTC and DAT training.
Based on empirical results, we adopt standard beam search for all models during inference. 
Further details are provided in Appendix~\ref{app:exp_setup}.

\paragraph{Evaluation.}
For translation quality, we adopt four commonly used metrics, which include two rule-based metrics, i.e., BLEU score\cite{bleu} and chrf~\cite{chrf}, and two model-based metrics, i.e., COMET~\cite{comet} and BLEURT~\cite{belurt}.
Specifically, for COMET, we utilize the wmt22-comet-da model \cite{comet22}, and for BLEURT, the BLEURT-20 model \cite{bleurt20} is employed.
~\citet{gpt4-eval} propose a GPT-based metric, namely GEMBA, to evaluate translation quality, and demonstrate state-of-the-art correlation with human labels.
We adopt GEMBA-GPT4-DA based on GPT-4~\cite{gpt4} as an advanced evaluation metric.
For human evaluation, we follow~\cite{mqm}, an evaluation methodology based on the Multidimensional Quality Metrics (MQM) framework, which provides a hierarchical analysis of translation errors.
Human evaluation details can be found in Appendix~\ref{app:human_eval}.

\section{Translation Quality}
\subsection{Automatic Evaluation}
The automatic evaluation results on WMT16 En$\Rightarrow$Ro and WMT21 De$\Rightarrow$En are presented in Table~\ref{tab:main_merge}. 
DAT obtains the most competitive performance compared with the AT counterpart across all automatic metrics, followed by MgMO and CTC.
MgMO and CTC achieve stronger performance than the representative iterative method, CMLM, when considering COMET and GEMBA which have shown better correlation with human annotation~\cite{comet,gpt4-eval}.
Notably, MgMO obtains comparable performance with CTC, without modifying model architecture.

\paragraph{Reliance on Knowledge Distillation.}
In both translation directions, all fully non-autoregressive methods except DAT and CMLM suffer more from training without distillation.
Typically, the vanilla NAT models suffer a decrease of more than 7 BLEU points without KD.
For strong NAT methods such as MgMO and CTC, on WMT21 De$\Rightarrow$En, the BLEU scores decrease by more than 2 and 3 points, respectively.
On the contrary, the performance of DAT and CMLM is as similarly affected as the AT counterpart, due to explicit dependency modeling similar to AT.
In the subsequent sections, we utilize knowledge distillation by default to analyze NAT models in the best-performing setting.

\paragraph{Evaluation Metrics.}
We consider a set of representative metrics to comprehensively compare NAT methods with AT.
We perform significance tests on all pairs of NAT models and their AT counterparts across all metrics.
Except for DAT, current NAT methods significantly underperform AT methods in various evaluation metrics including rule-based (BLEU and chrf), model-based (COMET and BLEURT), and GPT4-based metrics (GEMBA), particularly in the raw data setting.
A notable observation is that DAT models are more competitive with AT models when evaluated using rule-based metrics, which assess the similarity between generated text and references.
In contrast, AT models outperform DAT models significantly under model-based metrics or GPT4 evaluation (GEMBA).
These metrics evaluate translation quality by measuring semantic similarity between two sentences based on parametric knowledge.
To gain a deeper understanding of this phenomenon, we conduct human evaluation using a systematic and fine-grained framework, i.e., MQM~\cite{mqm}, to further compare NAT with AT.

\subsection{Human Evaluation}
		




\begin{table}[t!]
    \centering
    \setlength{\belowcaptionskip}{-0.2cm}
    \small
    \setlength{\tabcolsep}{4pt} 
    \begin{tabular}{lcccc}
        \toprule
        Model & MQM$\downarrow$ & FLC. Err$\downarrow$ & ACC. Err$\downarrow$ & NON. Err$\downarrow$  \\
        \midrule
        AT   & 176.67 & 34.33 & 142.00 & 0.33\\
        MgMO   & 301.33 & 52.00 & 240.00 & 9.33 \\
        CTC    & 360.67 & 63.33 & 280.67  & 16.67 \\
        DAT     & 229.33 & 38.00 & 183.67 &0.33 \\
        CMLM   & 375.67 & 47.33 & 153.33 & 175.00 \\
        \bottomrule
    \end{tabular}
    \caption{
    \label{tab:human_evaluation}
    Human evaluation results under MQM framework. MQM denotes weighted error counts of three major error types: fluency (FLC.), accuracy (ACC.) and non-translation (NON.). 
    }
    
\end{table}

\begin{figure}[t!]
\setlength{\belowcaptionskip}{-0.1cm}
\centering
\includegraphics[width=0.99\linewidth]{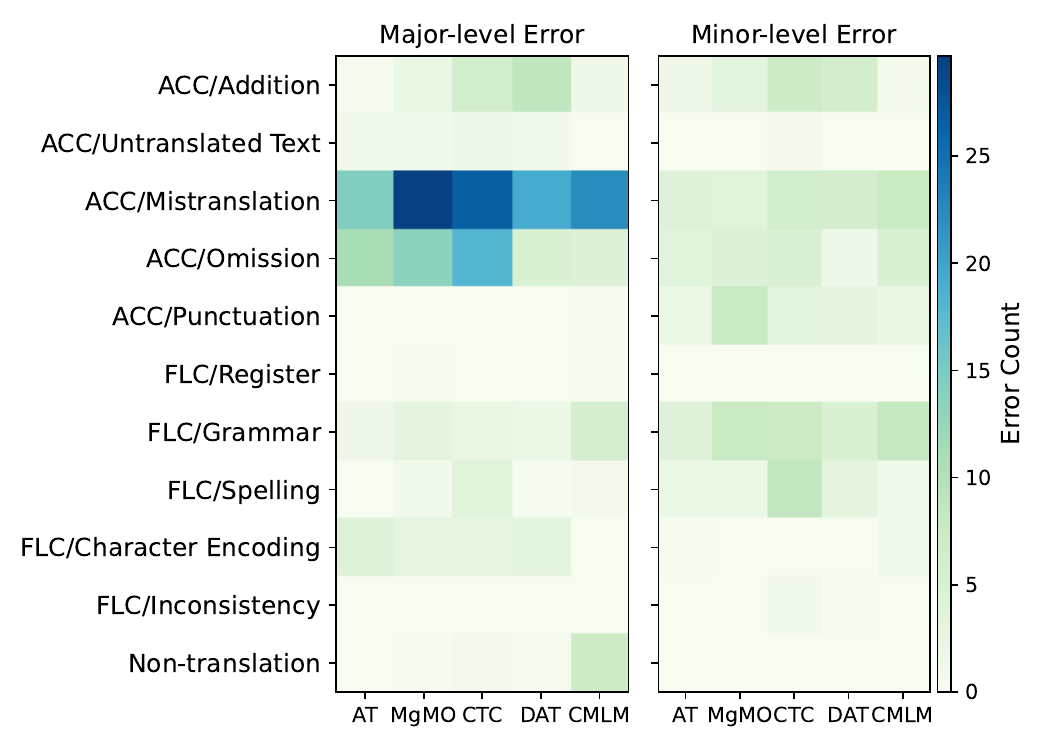}
\caption{
\label{fig:human_heat_map}
Heatmap visualization of MQM evaluation: darker colours indicate larger error counts for certain error types. The left side presents major-level errors while the right side shows minor-level errors.
}
\end{figure}

The evaluation results, obtained by averaging the error counts from three translators, are presented in Table~\ref{tab:human_evaluation}.
We omit human evaluation on the vanilla NAT due to its poor performance under automatic evaluation.
The performance ranking of human evaluation aligns with the automatic evaluation: AT performs the best, followed by DAT, MgMO, CTC, and CMLM.
Models with explicit dependency modeling (AT, DAT and CMLM) generate more fluent translations than those without (MgMO and CTC).
, with fewer fluency errors.
Despite comparable fluency to AT, DAT exhibits low translation accuracy.
All NAT methods, particularly CMLM, generate non-translations in certain cases.

A fine-grained error visualization is presented in Figure~\ref{fig:human_heat_map}.
The \textit{mistranslation} error type at the major level has the highest proportion among all models, with the models lacking explicit dependency (MgMO and CTC) producing the most errors.
AT performs generally better than NAT except for a considerable number of \textit{omission} errors. 
In contrast, NAT models tend to generate translations with additional or duplicated content (\textit{addition}), particularly CTC and DAT which increase decoder length to model latent alignments.
These two models also exhibit more \textit{spelling} errors.
Compared to AT, NAT models tend to produce more \textit{punctuation} errors and \textit{grammar} errors.
Similar to AT, MgMO and CTC translations also frequently lack partial source content (\textit{omission}).

We explore human annotations to understand typical patterns. 
Regarding \textit{omission} errors, AT often exhibits incomplete generation at the sentence's end. On the other hand, MgMO and CTC frequently omit content throughout the entire sentence, such as missing adjectives or verbs.
The NAT's \textit{grammar} errors primarily stem from incorrect verb tense and singular/plural usage, resulting from its limited language dependency modeling.
The case study indicates that, for CTC, the major \textit{addition} errors are attributed to generating words with \textit{spelling} errors, which are regarded as irrelevant content by annotators.
For DAT, these \textit{addition} errors stem from \textbf{n-gram repetition}, where the model generates a repeated segment from the previous context. 
For example, ``By \textit{the beginning of November}, there are seven races until \textit{the beginning of November}.''
To give an intuitive representation, we present several cases for the aforementioned error types in Appendix~\ref{app:case_study}.
All these patterns can be attributed to inadequate language dependency modeling with limited or redundant decoding length.

\subsection{Effects of Explicit Dependency}
\label{sec:explicit_dependency}
\paragraph{Repetition Ratio.}
We first examine token repetition ratio ~\cite{understanding_kd,axe,oaxe} in model translations, which is the ratio of generations with repeated tokens, e.g., ``He is is a lawyer''.
The results are shown in Table~\ref{tab:Repitition}.
We can observe that models without latent alignment modeling (NAT, MgMO and CMLM) suffer severe token repetition during generation.

		




\begin{table}[t!]
\setlength{\belowcaptionskip}{-0.2cm}
    \centering
    \small
    \setlength{\tabcolsep}{4pt} 
    \begin{tabular}{ccccccc}
        \toprule
        Ref. & AT & NAT & MgMO & CTC & DAT & CMLM \\
        \midrule
        0.00  & 0.50 & 27.64 & 16.47 & 1.85  & 0.00 & 14.52 \\
        \midrule
        0.00  & 0.10 & 31.10 & 23.50 & 1.60 & 0.00 & 12.60 \\
        \bottomrule
    \end{tabular}
    \caption{Uni-gram repetition ratios on WMT16 En$\Rightarrow$Ro (first row) and WMT21 De$\Rightarrow$En (second row). The term ``Ref.'' refers to the reference translation
    }
    \label{tab:Repitition}
\end{table}

\begin{figure}[t]
\setlength{\belowcaptionskip}{-0.1cm}
\centering
\includegraphics[width=0.8\linewidth]{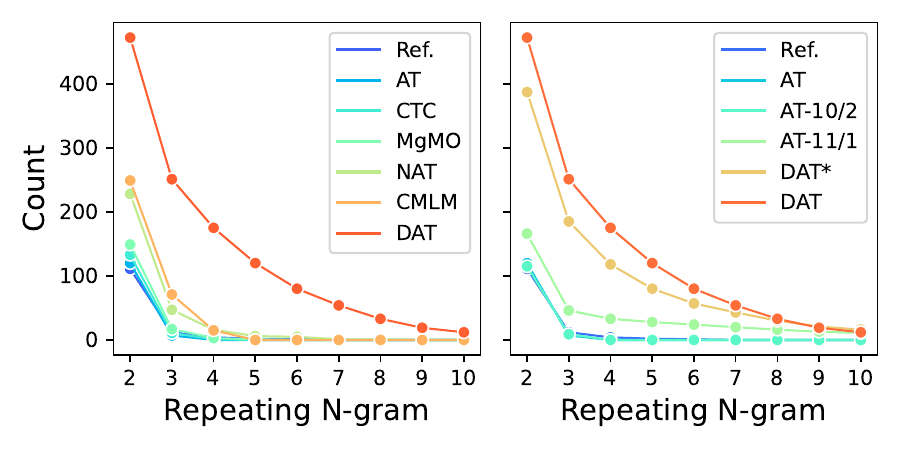}
\caption{
\label{fig:n-gram_repeat}
N-gram repetition of different models (WMT21 De$\Rightarrow$En), where the x-axis represents the size of the n-gram and the y-axis represents the count.
}
\end{figure}

\paragraph{N-gram Repetition.}
Besides consecutive uni-gram repetition, a more subtle phenomenon is nonadjacent n-gram repetition.
Such a repetition can be overlooked under traditional metrics such as BLEU score, which only calculates the n-gram precision of the generations.
Consequently, translations that contain n-gram repetition may even achieve higher BLEU scores.
This could explain why DAT performs better than AT under rule-based metrics but not under model-based or GPT4-based metrics.
We collect the n-gram repetition count for each model, as shown in the left part of Figure~\ref{fig:n-gram_repeat}.
We can observe that DAT demonstrates a stronger tendency to generate repeating n-grams with higher counts across various n-gram granularity (2 to 10), which aligns with a substantial number of \textit{addition} errors in human evaluation.

\paragraph{Enhancing Dependency Modeling.}
DAT utilizes one-linear-layer attention modules to model local vertex transitions.
Such explicit dependency modeling can be limited when dealing with long sequence generation.
For example, consider the sentence "By \textit{the beginning of November}, there are seven races until \textit{the beginning of November}." 
In this case, both the beginning and the end of the sentence are valid positions for the temporal prepositional phrase "the beginning of November"
In DAT, at the vertex state corresponding to the token "races," only information from that current vertex state is used to determine the index of the next vertex state using one-linear-layer attention layers.
In contrast, AT considers all previously generated tokens and utilizes Transformer decoder layers to determine the next token. 
Under weak dependency modeling in DAT, early generations can be ignored and repeated phrases can be falsely pointed to (e.g., ``the beginning of November'').
To validate this assumption, we train an asymmetrical AT model with a shallow decoder to simulate weak dependency modeling, and a deep encoder to guarantee model size.
As shown in Figure~\ref{fig:n-gram_repeat} (right-side), AT-11/1 (11-layer encoder and 1-layer decoder) also tends to generate repeated n-grams, and adding one decoder layer (AT-10/2) mitigates this issue.
Nevertheless, AT-11/1 performs better than DAT as it relies on the entire generation history rather than just considering the current token.
To alleviate this issue without influencing decoding efficiency, we introduce an additional linear layer for both $\mathbf{Q}$ and $\mathbf{K}$ to strengthen the token transition modeling.
This refined model is referred as DAT$^{*}$ (Appendix~\ref{app:dat_star}).
With 0.7\% additional parameters compared to DAT alone, we observe that DAT$^{*}$ exhibits less n-gram repetition while maintaining decoding speed.
We present a case study of how DAT$^*$ alleviates n-gram repetition in Appendix~\ref{app:case_study}.
Nevertheless, DAT$^{*}$ is limited in a one-step local transition foundation and cannot fundamentally resolve n-gram repetition.




\begin{figure}[t]
\setlength{\belowcaptionskip}{-0.1cm}
\centering
\includegraphics[width=0.65\linewidth]{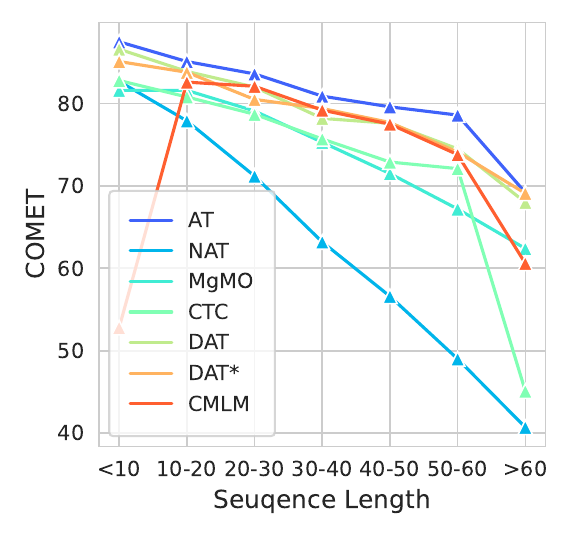}
\caption{
\label{fig:length}
Translation quality (COMET) w.r.t. source sequence length on WMT21 De$\Rightarrow$En.
}
\end{figure}

\section{Generalization and Robustness}

\paragraph{Length Generalization.}
Figure~\ref{fig:length} illustrates that all models experience a decline in performance as the length of the source sequence increases, albeit at varying rates.
AT surpasses all NAT methods in length generalization.
Models incorporating explicit dependency (e.g., AT, DAT, and CMLM) exhibit slower degradation compared to others.
Notably, CTC and CMLM experience severe performance drops on sequences longer than 60.

\begin{figure}[t!]
\setlength{\belowcaptionskip}{-0.1cm}
\centering
\includegraphics[width=0.8\linewidth]{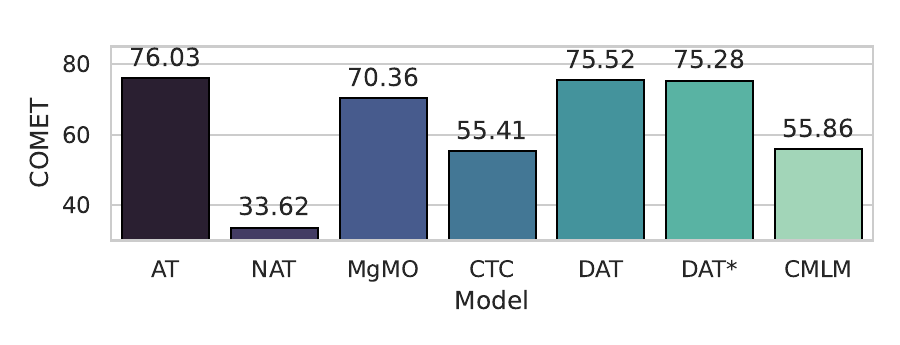}
\caption{
\label{fig:cross_domain}
Average cross-domain performance (COMET) of WMT21 De$\Rightarrow$En models on out-of-domain testsets.
}
\end{figure}

\paragraph{Cross-domain Generalization.}
Figure~\ref{fig:cross_domain} illustrates the cross-domain performance averaged across 5 domains.
Models with explicit dependency, such as AT and DAT, achieve high cross-domain performance.
On the other hand, CTC and CMLM demonstrate substantial degradation in performance when tested on out-of-domain datasets
This is due to CTC models generating spelling errors and CMLM models propagating errors from early steps.
These issues are further exacerbated in cross-domain testsets that contain more terminologies, leading to subpar performance.
The complete results can be found in Appendix~\ref{app:cd}.

\paragraph{Compositional Generalization.}


\begin{table}[t]
\setlength{\belowcaptionskip}{-0.1cm}
\centering
\small
\begin{tabular}{lccc}
\toprule
Model &    AT & DAT & DAT$^*$\\
\midrule
Instance-level CTER$\downarrow$  &  28.42\% &  43.66\% & 42.52\%   \\
Aggregate-level CTER$\downarrow$ &  62.88\% &   79.49\%  &79.12\%      \\
\bottomrule
\end{tabular}
\caption{Compositional generalization performance. \label{tab:cg}
}
\end{table}
We measure compositional generalization on GoGnition ~\cite{cg} which evaluates the ability to translate unseen phrases of simple and known semantic units.
The results are shown in Table~\ref{tab:cg} \footnote{We only evaluate AT and DAT as they do not rely on knowledge distillation.}.
Instance-level CTER and Aggregate-level CTER denote the compound translation error rates of translating novel compounds.
Despite the narrowing gap in in-domain and out-of-domain testsets, we observe a significant difference in compositional generalization between DAT and AT.
This discrepancy is reflected in higher error rates, indicating a disparity in dependency modeling capabilities.

\begin{figure}[t]
\setlength{\belowcaptionskip}{-0.1cm}
\centering
\includegraphics[width=0.8\linewidth]{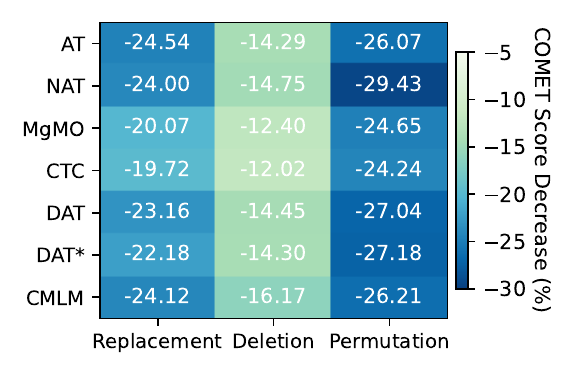}
\caption{
\label{fig:noise}
Translation performance (COMET) decreases (\%) on noisy testsets of WMT21 De$\Rightarrow$En, with darker colours indicating greater degradation.
}
\end{figure}

\paragraph{Robustness to Input Perturbations.}
Finally, we explore models' robustness to different input perturbations, including random replacement, deletion and permutation (Details in Appendix~\ref{app:exp_setup}), with results shown in Figure~\ref{fig:noise}.
In contrast to previous findings that suggest explicit modeling provides advantages,
models without explicit incorporation of modeling (e.g., MgMO and CTC) are less affected by input noises.
This can be because explicit dependency generation may introduce exposure bias~\cite{exposure1,exposure2}, where errors occurring at early time steps (AT and DAT) or iterative steps (CMLM) can accumulate and propagate into future predictions, making them susceptible to input perturbations.
For complete results, please refer to Appendix~\ref{app:noise}.

To the best of our knowledge, this is the first comparison of NAT and AT in terms of generalization and robustness.
In addition to the disparity in translation performance on benchmark datasets, inadequate language dependency modeling causes NAT methods to significantly lag behind AT.
However, this weak dependency does provide an advantage in resisting input perturbations.

\section{Related Work}
We discuss several representative NAT methods in Section~\ref{sec:method}. 
A more detailed discussion on NAT advances is presented in Appendix~\ref{app:advance_nat}.
Different from surveys~\cite{survey1,survey2} that conduct a comprehensive survey on recent NAT advances, we focus on comparing NAT with AT comprehensively.
Our work is also related to previous work analyzing neural machine translation (Appendix~\ref{app:nmt_analysis}).
\citet{understanding_kd} find that knowledge distillation boosts NAT performance by reducing data complexity.
\citet{multilingual_nat} discuss knowledge transfer in the context of multilingual NAT.
~\citet{leaning_nat} understand the learning process of NAT both theoretically and empirically.
Differently, we focus on systematically comparing common NAT techniques with their AT counterparts in a systematic manner to showcase existing performance gaps for future research.

\section{Conclusion}
We compared representative NAT methods with AT under a comprehensive evaluation that encompasses a set of evaluation dimensions, including human evaluation.
Our research aims to fill in the research gap of the real competitiveness of NAT to AT.
Both automatic and human evaluations indicated that despite the narrowing gap, NAT methods underperform AT, with varying error patterns such as translation omission, spelling errors and n-gram repetitions.
Our empirical results and analyses demonstrated that explicit dependency modeling is crucial for generating human-like languages, although strong dependence can suffer explore bias.
Future research on NAT should focus on how to consolidate explicit language dependency while maintaining decoding efficiency.

\section*{Limitations}
We systematically evaluate NAT and AT, highlighting performance gaps for future research. However, there are limitations:
Firstly, we assess state-of-the-art NAT models using research-oriented datasets (WMT, OOD, CG), which mainly consist of English-centric text with a formal style and limited topic range.
Secondly, each NAT model is annotated with only 100 samples. This may not cover all potential error types.
Finally, we focus primarily on fully non-autoregressive methods due to their superior decoding efficiency. Our results are also limited to training-from-scratch methods; extending conclusions to large language models is left for future work.

\section*{Ethical Considerations}
We honor the ACL Code of Ethics. 
No private data or non-public information is used in this work.
For human annotation, we hired three annotators who have degrees in English Linguistics or Applied Linguistics. 
Before formal annotation, annotators were asked to annotate $100$ samples randomly extracted from the dataset, and based on average annotation time we set a fair salary (i.e., 32 dollars per hour) for them. 
During their training annotation process, they were paid as well. 
The annotation does not involve any personally sensitive information. 
The annotation strictly follows the annotation guide of MQM~\cite{mqm}, with details presented in Appendix~\ref{app:human_eval}.
We adhere to the terms of companies offering commercial LLM APIs and express our gratitude to all global collaborators for their assistance in utilizing these APIs.

\section*{Acknowledgement}
We would like to thank all reviewers for their insightful comments and suggestions to help improve the paper. 
This publication has emanated from research conducted with the financial support of the Ministry of Science and Technology of China under Grant Number 2022YFE0204900.
This work is also supported by a grant from Lan-bridge Information Technology Co., Ltd. 
We extend our gratitude to colleagues at Lan-bridge for their professional annotations and evaluations of model translations. Special thanks to Xianchao Zhu and Jing Li for their invaluable assistance in organizing the annotation process.

\bibliography{acl}

\clearpage
\appendix

\section{Recent Advances in NAT}
\label{app:advance_nat}
Various techniques have been proposed to address the performance limitations of NAT.
\citet{clnat,tclnat,mgclnat,glat} devise dedicated training curriculums to reduce the learning difficulty of NAT models, whereas \citet{kdnat,lexinat,lowfwords} propose improved distillation training.
Latent variable modeling has received significant attention in enhancing NAT performance \cite{ctc18,dislatent,flowseq,ctc,latentglat,tgtcodenat}.
Typically, \citet{da} explicitly models target dependency as paths in a directed acyclic graph.
Another line of research focuses on enhancing the cross-entropy loss or alternating to metric-based objectives \cite{natcrf,natbow,axe,oaxe,bleunat,mgmo}.
In contrast to fully non-autoregressive methods mentioned earlier, another approach decomposes one-shot generation into multiple iterative non-autoregressive generations ~\cite{levnat,maskp,cmlmc}.
\citet{clarity} align common NAT techniques and compare translation quality and speed implications under uniform environments.
Despite claiming improved performance and comparability with autoregressive models (AT), these approaches are limited in their evaluation using rule-based metrics like BLEU score~\cite{bleu}, which demonstrates poor correlation with human preference~\cite{comet,stop-bleu}.

\section{Analysis Research in NMT}
\label{app:nmt_analysis}
\citet{lens_smt} interpret NMT's learning process during training, and
~\citet{black_box,digging_error} interpret and analyze model predictions during inference.
~\citet{domain_robust} study NMT generalization ability to novel domains, whereas ~\citet{cg} demonstrate that NMT's weak compositional generalization capability.
Additional metrics proposed by ~\citet{input_robust}  quantify the effects of input perturbations.
Hallucination, which refers to the generation of unrelated outputs by the model, has also been extensively studied  ~\cite{understand_hallu,haystack}.


\section{Experiment Setup}
\label{app:exp_setup}
\paragraph{Datasets}
To evaluate general translation performance, we choose WMT16 En$\Rightarrow$Ro, a widely used benchmark dataset for non-autoregressive translation.
In addition, we select a large-scale benchmark dataset, i.e., WMT21 De$\Rightarrow$En, which consists of 101.35M parallel sentences and is further filtered to 88.66M.
We apply BPE~\cite{bpe} on the concatenated training sets with 32,000 operations.
Knowledge distillation is commonly used for training NAT models \cite{nat, natcrf, maskp, axe}. 
We train an autoregressive Transformer base model on the raw dataset as the teacher model and use it to generate the distilled dataset.
To assess cross-domain translation, we employ the test sets from ~\citep{cross-domain}, which encompass test instances from 5 domains: medical, IT, koran, law, and subtitles, and we evaluate the models (trained on WMT21 De$\Leftrightarrow$En) on these test sets.
For compositional generalization, we utilize CoGnition~\cite{cg} with its original data configurations.
Following the approach in~\cite{mt-scale}, we measure model robustness on the WMT21 De$\Rightarrow$En testset by introducing three types of input noise: (1) word deletion with a probability of 0.1; (2) word replacement with "<unk>" with a probability of 0.1; (3) word swapping within a range of 3 words with a probability of 0.1.

\begin{table*}[t!]
\setlength{\tabcolsep}{2pt}
    \centering
    \small
    \setlength{\belowcaptionskip}{-0.2cm}
\begin{tabular}{lccccccccccc}
\toprule
Model & BLEU & COMET & 2-gram &  3-gram & 4-gram & 5-gram & 6-gram & 7-gram & 8-gram & 9-gram & 10-gram \\
\midrule
AT & 32.04 & 84.72 & 130 & 10 & 0 & 0 & 0 & 0 & 0 & 0 & 0 \\
DAT(Beam) & 32.26 & 83.32 & 472 & 251 & 175 & 120 & 80 & 54 & 33 & 19 & 12 \\
DAT(Viterbi) & 31.99 & 82.41 & 410 & 193 & 126 & 78 & 53 & 34 & 23 & 17 & 6 \\
MgMO & 30.32 & 81.15 & 149 & 17 & 3 & 1 & 0 & 0 & 0 & 0 & 0 \\
CTC & 30.52 & 80.06 & 133 & 11 & 1 & 0 & 0 & 0 & 0 & 0 & 0 \\
\bottomrule
\end{tabular}

    \caption{
    \label{tab:viterbi}
    Comparison of different models on WMT21 De$\Rightarrow$En translation task with distillation. The table reports BLEU, COMET, and N-gram repetition rates (2-grams to 10-grams) for each model. DAT(Beam) and DAT(Viterbi) are two variants of the DAT model.
    }
    
\end{table*}

\paragraph{Model Settings}
We adhere to the best-performing model configuration outlined in the corresponding papers~\cite{VaswaniSPUJGKP17,nat,ctc,mgmo,da,maskp}.
For all models, we utilize Transformer with a Transformer\_Base configuration: both the encoder and decoder comprise 6 layers with 8 attention heads. The hidden dimension is set to 512, while the feedforward layer dimension is set to 2,048.
The model is trained using Adam \cite{adam} optimizer.
We apply a weight decay of 0.01 and label smoothing of 0.1.
The learning rate initially increases to $5\cdot10^{-4}$ within the first 10K steps and subsequently decays exponentially. For glancing training, the glancing probability is gradually annealed from 0.5 to 0.1 in 200k steps.
All results are based on models trained on the KD dataset unless otherwise stated.
For inference, we present results obtained through beam search with a beam size of 5. 
In the case of iterative models such as CMLM, we set the number of iterative steps as 10.
Following the official DA-Transformer guidelines\footnote{\url{https://github.com/thu-coai/DA-Transformer/tree/v1.0}}, we compared Viterbi Decoding with beam search and found that beam search slightly outperformed Viterbi in terms of BLEU and COMET metrics. Therefore, we chose beam search to present DAT's best performance.
The performance comparison, including BLEU, COMET and n-gram repetition is presented in the following table~\ref{tab:viterbi}. Despite Viterbi Decoding's reduction in n-gram repetition, this issue was still more pronounced in DAT compared to AT and other NAT methods, aligning with the conclusion in our paper. 
We utilized 4 NVIDIA V100 GPUs for our computations, dedicating two days for the CTC process and five days for DA. Other methods were executed within one day each.

\section{Human Annotation}
\label{app:human_eval}
We follow~\citet{mqm}, an evaluation methodology based on the Multidimensional Quality Metrics (MQM) framework, which provides a hierarchical analysis of translation errors.
We adopt two common error hierarchy categories: \textit{Accuracy} and \textit{Fluency}. \textit{Accuracy} covers fine-grained 4 error sub-types such as \textit{Addition}, \textit{Omission}, \textit{Mistranslation} and \textit{Untranslated Text}, whereas \textit{Fluency} covers \textit{Punctuation}, \textit{Spelling}, \textit{Grammar}, \textit{Register}, \textit{Inconsistency} and \textit{Character Encoding}.
Translations that are too badly garbled to permit error classification are classified as \textit{Non-translation}.
In addition to the error type, each error is also annotated with a severity label: minor and major.
We follow the error weighting in~\citet{mqm} to compute the weighted error counts for each system.
Annotation details are presented in Appendix~\ref{app_human}.
We hire three expert translators to conduct side-by-side human evaluations on the 5 German-English translation models, i.e., AT, NAT, MgMO, CTC, DA and CMLM.
We randomly sample 100 translations from the WMT21 De$\Rightarrow$En testset and ask translators to annotate translation errors for each instance following the MQM annotation guideline.
We average the error counts from the 3 annotators as human evaluation results.
For conducting human annotation, we hired three annotators who have degrees in English Linguistics or Applied Linguistics. 
Before formal annotation, annotators were asked to annotate $100$ sampled translations from 5 systems, and based on average annotation time we set a fair salary (i.e., 32 dollars per hour) for them. 
During their training annotation process, they were paid as well. 
\section{MQM Annotation}
\label{app_human}
We present the details of the error type description in Table~\ref{tab:mqm-hierarchy}, the error severity description in Table~\ref{tab:mqm-severity} and error weights in Table~\ref{tab:mqm-scoring}.
\begin{table}[t]\centering
\scalebox{0.80}{
\begin{tabular}{lll}\toprule
Severity & Category & Weight \\
\midrule
Major & Non-translation & 25 \\
      & all others & 5 \\
\midrule
Minor & Fluency/Punctuation & 0.1 \\
      & all others          & 1 \\
\bottomrule
\end{tabular}
}
\caption{MQM error weighting~\cite{mqm}.}
\label{tab:mqm-scoring}
\vspace{-1em}
\end{table}

\begin{table*}[htb]\centering
\scalebox{0.80}{
\begin{tabular}{lll}\toprule
\multicolumn{2}{l}{Error Category} & Description \\
\midrule
Accuracy & Addition    & Translation includes information not present in the source or repeated content. \\
    & Omission         & Translation is missing content from the source. \\
    & Mistranslation   & Translation does not accurately represent the source.\\
    & Untranslated text & Source text has been left untranslated. \\
\midrule
Fluency & Punctuation   & Incorrect punctuation (for locale or style). \\
    & Spelling          & Incorrect spelling or capitalization. \\
    & Grammar           & Problems with grammar, other than orthography. \\
    & Register          & Wrong grammatical register (eg, inappropriately informal pronouns). \\
    & Inconsistency     & Internal inconsistency (not related to terminology). \\
    & Character encoding          & Characters are garbled due to incorrect encoding. \\
\midrule
Non-translation & & Impossible to reliably characterize the 5 most severe errors.\\
\bottomrule
\multicolumn{3}{c}{}\\
\end{tabular}
}
\caption{MQM hierarchy~\cite{mqm}.}
\label{tab:mqm-hierarchy}
\end{table*}

\begin{table*}[!htb]\centering
\scalebox{0.80}{
\begin{tabular}{p{0.09\textwidth}p{\textwidth}}\toprule
Severity & Description \\
\midrule
Major & Errors that may confuse or mislead the reader due to significant change in meaning or because they appear in a visible or important part of the content. \\
\midrule
Minor & Errors that don't lead to loss of meaning and wouldn't confuse or mislead the reader but would be noticed, would decrease stylistic quality, fluency or clarity, or would make the content less appealing.\\
\bottomrule
\end{tabular}
}
\caption{MQM severity levels~\cite{mqm}.}
\label{tab:mqm-severity}
\end{table*}

\section{Human Evaluation Results}
The annotation results (average from 3 translators) are presented in Table~\ref{tab:human_evaluation_detail_div3}.

\begin{table*}[t]
    \centering
    \small
    \begin{tabular}{lcccccccccc}
        \toprule
        & \multicolumn{2}{c}{AT} & \multicolumn{2}{c}{MGMO} & \multicolumn{2}{c}{CTC} & \multicolumn{2}{c}{DA} & \multicolumn{2}{c}{CMLM} \\
        \cmidrule(lr){2-3} \cmidrule(lr){4-5} \cmidrule(lr){6-7} \cmidrule(lr){8-9} \cmidrule(lr){10-11}
        & \textbf{Maj.} & \textbf{Min.} & \textbf{Maj.} & \textbf{Min.} & \textbf{Maj.} & \textbf{Min.} & \textbf{Maj.} & \textbf{Min.} & \textbf{Maj.} & \textbf{Min.} \\
        \midrule
        ACC/Addition & 0.33 & 1.33 & 2.00 & 3.33 & 6.33 & 7.00 & 8.67 & 6.00 & 1.33 & 0.67 \\
        ACC/Untranslated Text & 1.00 & 0 & 1.00 & 0.00 & 1.33 & 0.67 & 1.00 & 0.00 & 0.00 & 0.00 \\
        ACC/Mistranslation & 14.33 & 4.00 & 29.67 & 3.67 & 26.67 & 6.33 & 19.33 & 6.00 & 22.33 & 7.67 \\
        ACC/Omission & 11.00 & 3.67 & 13.33 & 4.33 & 18.00 & 4.67 & 5.00 & 1.33 & 4.33 & 5.00 \\
        ACC/Punctuation & 0.00 & 1.67 & 0.00 & 7.67 & 0.00 & 3.33 & 0.00 & 3.00 & 0.33 & 2.33 \\
        FLC/Register & 0.00 & 0.00 & 0.33 & 0.00 & 0.00 & 0.00 & 0.00 & 0.00 & 0.33 & 0.00 \\
        FLC/Grammar & 1.33 & 4.00 & 3.00 & 7.67 & 2.33 & 7.33 & 1.67 & 5.00 & 5.67 & 8.00 \\
        FLC/Spelling & 0.00 & 1.67 & 1.00 & 1.67 & 3.67 & 8.33 & 0.33 & 3.00 & 0.67 & 1.00 \\
        FLC/Character Encoding & 4.00 & 0.33 & 2.67 & 0.00 & 2.67 & 0.00 & 3.33 & 0.00 & 0.00 & 1.00 \\
        FLC/Inconsistency & 0.00 & 0.00 & 0.00 & 0.00 & 0.00 & 1.00 & 0.00 & 0.33 & 0.00 & 0.00 \\
        Non-Translation & 0.00 & 0.00 & 0.33 & 0.00 & 0.67 & 0.00 & 0.33 & 0.00 & 7.00 & 0.00 \\
        \bottomrule
    \end{tabular}
    \caption{Human Evaluation Results - Error Counts by Type (Averaged from Three Translators' Annotations).}
    \label{tab:human_evaluation_detail_div3}
\end{table*}


\section{Case Study}
\label{app:case_study}

We present a case study to showcase the n-gram repetition phenomenon in Table~\ref{tab:case_study}.
We present several cases to showcase the \textit{spelling} errors of CTC and DAT in Table~\ref{tab:case_study_ctc}.
A case study of \textit{omission} errors is shown in Table~\ref{tab:case_study_omission}.
A case study of \textit{grammar} and \textit{punctuation} errors is shown in Table~\ref{tab:case_study_grammar}.
A case study of how DAT$^*$ alleviates n-gram repetition is presented in Table~\ref{tab:dat_vertex_case}.
\begin{table*}[t!]
    \setlength{\belowcaptionskip}{-0.cm}
    \centering
    \small
    \renewcommand{\arraystretch}{1.2} 
    \begin{tabular}{lp{0.76\linewidth}} 
    \toprule
    \multicolumn{2}{c}{\textbf{Case 1}} \\
    \midrule
      \textbf{Source Sentence} & In Sachen Kindergarten- respektive Krippenplätzen hat sie bereits Kontakt mit einer örtlichen Einrichtung aufgenommen.  \\
      \midrule
      \textbf{Reference Sentence} & Regarding kindergarten respectively nursery places she has already established contact with the local facilities. \\
      \midrule
      \textbf{DAT Translation} &  \colorbox{highlightcolor}{She has already made contact with a local institution} in terms of kindergarten and crib places, \colorbox{highlightcolor}{she has already made contact with a local institution}. \\
      \midrule
      \textbf{DAT$^{*}$ Translation} & In terms of kindergarten or crib places, she has already contacted a local institution. \\
    \midrule
    \multicolumn{2}{c}{\textbf{Case 2}} \\
    \midrule
      \textbf{Source Sentence} &  31 Spieler begrüßte er an der Säbener Straße, darunter auch die neuen Akteure um Edel-Einkauf Leroy Sané, der erstmals nach seinem Wechsel von Manchester City alle neuen Kollegen auf dem Platz traf.
 \\
      \midrule
      \textbf{Reference Sentence} & He greeted 31 players at the Säbener Straße, among them the new players around special purchase Leroy Sané who met all new colleagues on the field for the first time after his transfer from Manchester City.
  \\
      \midrule
      \textbf{DAT Translation} & He welcomed 31 players on Säbener Straße, including the new players for fine shopping Leroy Sané, who met all new colleagues on the square \colorbox{highlightcolor}{for the first time after his move from Manchester City}, met all the new colleagues on the pitch \colorbox{highlightcolor}{for the first time after his move from Manchester City}.
  \\
      \midrule
      \textbf{DAT$^{*}$ Translation} & He welcomed 31 players on Säbener Straße, including the new players around fine shopping Leroy Sané, who met all new colleagues on the pitch for the first time after his move from Manchester City.
  \\
    \midrule
    \multicolumn{2}{c}{\textbf{Case 3}} \\
    \midrule
      \textbf{Source Sentence} & In der Stadt Oakland in Kalifornien wurde ein Gerichtsgebäude in Brand gesteckt.  \\
      \midrule
      \textbf{Reference Sentence} &  A courthouse was set on fire in Oakland, California. \\
      \midrule
      \textbf{DAT Translation} & \colorbox{highlightcolor}{In the city of Oakland, California}, a courthouse was set on fire \colorbox{highlightcolor}{in the city of Oakland, California}.\\
      \midrule
      \textbf{DAT$^{*}$ Translation} &  A courthouse was set on fire in the city of Oakland, California.\\
    \midrule
    \multicolumn{2}{c}{\textbf{Case 4}} \\
    \midrule
      \textbf{Source Sentence} & Die Windkraftwerke auf der deutschen Nordsee haben in den ersten sechs Monaten des Jahres 11,51 Terawattstunden Strom in das Netz eingespeist.  \\
      \midrule
      \textbf{Reference Sentence} &  The wind power plants of the German North Sea delivered 11.51 terawatt hours electricity to the net in the first six months of the year. \\
      \midrule
      \textbf{DAT Translation} & \colorbox{highlightcolor}{In the first six months of the year}, the wind power plants on the German North Sea fed 11.51 terawatt hours of electricity into the grid \colorbox{highlightcolor}{in the first six months of the year}. \\
      \midrule
      \textbf{DAT$^{*}$ Translation} &  The wind power plants on the German North Sea fed 11.51 terawatt hours of electricity into the grid in the first six months of the year. \\
   \midrule
    \multicolumn{2}{c}{\textbf{Case 5}} \\
    \midrule
      \textbf{Source Sentence} & Bis Anfang November stehen sieben Rennen an.  \\
      \midrule
      \textbf{Reference Sentence} &  Until the beginning of November seven races are planned. \\
      \midrule
      \textbf{DAT Translation} & By \colorbox{highlightcolor}{the beginning of November}, there are seven races until \colorbox{highlightcolor}{the beginning of November}. \\
      \midrule
      \textbf{DAT$^{*}$ Translation} &  There are seven races until the beginning of November. \\

    \bottomrule
    \end{tabular}
    \caption{A case study of n-gram repeating of DAT models, comparing with DAT$^*$ which enhances dependency modeling by adding a linear layer.
    The text in the grey background denotes the repeated segment.
    }
    \label{tab:case_study}
\end{table*}

\begin{table*}[t!]
    \setlength{\belowcaptionskip}{-0.cm}
    \centering
    \small
    \renewcommand{\arraystretch}{1.5} 
    \begin{tabular}{lp{0.7\linewidth}} 
    \toprule
    \multicolumn{2}{c}{\textbf{Case 1}} \\
    \midrule
      \textbf{Source Sentence} & Sie steckten vor einem Jugendgefängnis Bauwagen in Brand, die Polizei setzte Blendgranaten und Pfefferspray ein. 
  \\
      \midrule
      \textbf{Reference Sentence} & They set construction trailers on fire in front of a youth detention center, the police used stun grenades and pepper spray.\\
      \midrule
      \textbf{CTC Translation} &  They set construction fire in front of a youth prison, the police used \colorbox{highlightcolor}{glgrenades}(glare grenades) and pepper spray. 
\\
     \midrule
  \multicolumn{2}{c}{\textbf{Case 2}} \\
    \midrule
      \textbf{Source Sentence} & Es gebe aber keine Anhaltspunkte, dass die Anzahl von illegalen Autorennen tatsächlich steige. \\
      \midrule
      \textbf{Reference Sentence} & However, there are no real indications that the number of illegal car races does in fact increase.\\
      \midrule
      \textbf{CTC Translation} &  However, there is no \colorbox{highlightcolor}{indicevidence}(indication/evidence) that the number of illegal car races is actually increasing. 
\\
     \midrule
  \multicolumn{2}{c}{\textbf{Case 3}} \\
    \midrule
      \textbf{Source Sentence} & Auf der A81 registriert die Polizei sogar mehr Rennen als auf jeder anderen Bundesautobahn.  
  \\
      \midrule
      \textbf{Reference Sentence} & The police registers even more races on the A81 than on any other federal autobahn. \\
      \midrule
      \textbf{CTC Translation} &  On the A81, the police \colorbox{highlightcolor}{registregister}(register) even more races than on any other federal highway.  \\
     \midrule
 \multicolumn{2}{c}{\textbf{Case 4}} \\
    \midrule
      \textbf{Source Sentence} & Im 24-Stunden-Vergleich wurden in Wien 60 Corona-Neuinfektionen gemeldet - in Niederösterreich gab es 22 Neuinfektionen.  \\
      \midrule
      \textbf{Reference Sentence} & In a 24 hour comparison 60 Corona new infections were reported in Vienna - in Lower Austria there were 22 new infections.\\
      \midrule
      \textbf{DAT Translation} & In a 24-hour comparison, 60 \colorbox{highlightcolor}{corona}(Corona) new infections were reported in Vienna - in Lower Austria there were 22 new infections.      \\
     \midrule
  \multicolumn{2}{c}{\textbf{Case 5}} \\
    \midrule
      \textbf{Source Sentence} & "Ich denke es ist uns gelungen, Rakoczy-Flair zu verbreiten", sagt Kurdirektorin Sylvie Thormann.  \\
      \midrule
      \textbf{Reference Sentence} & “I think we still succeeded in spreading Rakoczy flair,” said the Kurstadt director, Sylvie Thormann. \\
      \midrule
      \textbf{DAT Translation} & “I think we have succeeded in spreading \colorbox{highlightcolor}{rakoczy}(Rakoczy) flair,“ says Prime Director Sylvie Thormann   
   \\
    \bottomrule
    \end{tabular}
    \caption{
    \label{tab:case_study_ctc}
    A case study of spelling errors of CTC and DAT. The text in the gray background indicates segments with spelling errors, followed by the correct spelling enclosed in brackets.
    }
    
\end{table*}

\begin{table*}[t!]
    \setlength{\belowcaptionskip}{-0.cm}
    \centering
    \small
    \renewcommand{\arraystretch}{1.5} 
    \begin{tabular}{lp{0.7\linewidth}} 
    \toprule
    \multicolumn{2}{c}{\textbf{Case 1}} \\
    \midrule
      \textbf{Source Sentence} & In diesem Jahr sind die Fluten besonders schlimm, was Wissenschaflter auf den Klimawandel zurückführen. 
  \\
      \midrule
      \textbf{Reference Sentence} & The floods were especially bad this year, which scientists have connected to climate change.\\
      \midrule
      \textbf{CTC Translation} &  This year, the floods are particularly bad, which scientists (have connected) to climate change.
\\
     \midrule
  \multicolumn{2}{c}{\textbf{Case 2}} \\
    \midrule
      \textbf{Source Sentence} & Den Punkterekord im englischen Fußball verpasste Coach Jürgen Klopp mit seinem Team nur knapp. \\
      \midrule
      \textbf{Reference Sentence} & Coach Jürgen Klopp with his team only narrowly missed the points record in English soccer.\\
      \midrule
      \textbf{CTC Translation} &  Coach Jürgen Klopp narrowly missed the (points) and his team in English football.	 
\\
 \midrule
  \multicolumn{2}{c}{\textbf{Case 3}} \\
    \midrule
      \textbf{Source Sentence} & Zuletzt hatten Thole/Wickler im September des vergangenen Jahres beim World Tour Final in Rom gespielt.  \\
      \midrule
      \textbf{Reference Sentence} & Thole/Wickler recently played in the World Tour Final in Rome in September of last year.\\
      \midrule
      \textbf{CTC Translation} & Thole/Wickler last (year) played at the World Tour Final in Rome (in) September. 
   \\
     \midrule
  \multicolumn{2}{c}{\textbf{Case 4}} \\
    \midrule
      \textbf{Source Sentence} & Acht Filme drehte sie mit dem Herzensbrecher.  
  \\
      \midrule
      \textbf{Reference Sentence} & She filmed eight films with the heart breaker. \\
      \midrule
      \textbf{MgMO Translation} &  She filmed eight films with the (heart) breaker.  \\
     \midrule
 
 \multicolumn{2}{c}{\textbf{Case 5}} \\
    \midrule
      \textbf{Source Sentence} & Frankfurt/Main - Der siebenmalige Zeitfahrweltmeister Tony Martin kann sich durchaus vorstellen, seine Radsport-Karriere fortzusetzen.  \\
      \midrule
      \textbf{Reference Sentence} & Frankfurt/Main - The seven-time time trial specialist Tony Martin can clearly picture continuing his bicycling career.\\
      \midrule
      \textbf{MgMO Translation} & Frankfurt/Main - The seven-time time-trial world champion Tony Martin can (clearly) imagine continuing his cycling career.	      \\
    
    \bottomrule
    \end{tabular}
    \caption{
    \label{tab:case_study_omission}
    A case study of omission errors of CTC and MgMO. The text indicated within brackets highlights the segments missed by models. 
    }
    
\end{table*}

\begin{table*}[t!]
    \setlength{\belowcaptionskip}{-0.cm}
    \centering
    \small
    \renewcommand{\arraystretch}{1.5} 
    \begin{tabular}{lp{0.7\linewidth}} 
    \toprule
    \multicolumn{2}{c}{\textbf{Case 1}} \\
    \midrule
      \textbf{Source Sentence} & Auf der A81 registriert die Polizei sogar mehr Rennen als auf jeder anderen Bundesautobahn. 
  \\
      \midrule
      \textbf{Reference Sentence} & The police register even more races on the A81 than on any other federal autobahn.\\
      \midrule
      \textbf{DA Translation} &  On the A81, the police \colorbox{highlightcolor}{registered} (register) even more races than on any other federal motorway.	
\\
     \midrule
  \multicolumn{2}{c}{\textbf{Case 2}} \\
    \midrule
      \textbf{Source Sentence} & Auch in der amerikanischen Metropole Seattle lieferten sich Demonstranten am Samstag Zusammenstöße mit der Polizei. \\
      \midrule
      \textbf{Reference Sentence} & In the American metropolis of Seattle demonstrators also ran into clashes with police on Saturday. \\
      \midrule
      \textbf{CTC Translation} &  In the American metropolis of Seattle, demonstrators also \colorbox{highlightcolor}{clashes} (clash/clashed) with the police on Saturday.\\
      \midrule
      \textbf{MgMO Translation} &  In the American metropolis of Seattle, demonstrators also \colorbox{highlightcolor}{clashes} (clash/clashed) with the police on Saturday.
      
\\
 \midrule
  \multicolumn{2}{c}{\textbf{Case 3}} \\
    \midrule
      \textbf{Source Sentence} & Die Polizei war seit dem frühen Abend mit zahlreichen Beamten im Einsatz, im gesamten Stadtgebiet war ein größeres Polizeiaufgebot zu sehen. \\
      \midrule
      \textbf{Reference Sentence} & The police was in use with numerous officers since the early evening, a major police detachment was observed in the entire city area.\\
      \midrule
      \textbf{CMLM Translation} & The police \colorbox{highlightcolor}{have been working} (worked) since the early evening with numerous officials, with a larger police squad throughout the city. 
   \\
     \midrule
  \multicolumn{2}{c}{\textbf{Case 4}} \\
    \midrule
      \textbf{Source Sentence} & Zuletzt hielten sich noch einige Dutzend Menschen auf dem Platz auf, verließen ihn jedoch vor Beginn der Sperrstunde um 1 Uhr.  
  \\
      \midrule
      \textbf{Reference Sentence} &Until last, some dozens of people were still present at the place, however, they also left before beginning of the curfew at 1 a.m. \\
      \midrule
      \textbf{CTC Translation} &  Finally, a few dozen people stayed on the square, but left it before the start of the curfew at 1 \colorbox{highlightcolor}{o clock} (o'clock).  \\
     \midrule
 
 \multicolumn{2}{c}{\textbf{Case 5}} \\
    \midrule
      \textbf{Source Sentence} & Wie die Polizei mitteilt, kam es danach wieder zu Auseinandersetzungen zwischen den beiden Personen.  \\
      \midrule
      \textbf{Reference Sentence} & Another scuffle followed between the two persons, according to the police.\\
      \midrule
      \textbf{MgMO Translation} & As the police say, there were clashes between the two people \colorbox{highlightcolor}{}(.)    \\
    
    \bottomrule
    \end{tabular}
    \caption{
    \label{tab:case_study_grammar}
    A case study of grammar and punctuation errors. The text in the gray background indicates segments with errors, followed by the correct format enclosed in brackets.
    }
    
\end{table*}

\begin{table*}[t!]
    \setlength{\belowcaptionskip}{-0.cm}
    \centering
    \small
    \renewcommand{\arraystretch}{1.5} 
    \begin{tabular}{lp{0.7\linewidth}} 
    \toprule
    \multicolumn{2}{c}{\textbf{Case 1}} \\
    \midrule
      \textbf{Source Sentence} & Bis Anfang November stehen sieben Rennen an.  \\
      \midrule
      \textbf{Reference Sentence} &  Until the beginning of November seven races are planned. \\
      \midrule
      \textbf{DAT Vertex Predictions} &  <BOS> By The Seven There Seven races are been races By \textbf{the} As \textbf{of} early \textbf{beginning} \textbf{of} \textbf{the} \textbf{beginning} \textbf{of} \textbf{of} \textbf{November} , there there will there will be be been are been are seven seven event@@ seven races races have seven seven the races p@@ races are place have run are been planned in p@@ ending scheduled until \textbf{the} run scheduled p@@ ending up by \textbf{beginning} \textbf{the} early \textbf{beginning} \textbf{beginning} \textbf{of} early \textbf{of} \textbf{November} \textbf{November} \textbf{beginning} \textbf{November} . <EOS> \\
      \midrule
      \textbf{DAT Translation} & By \colorbox{highlightcolor}{the beginning of November}, there are seven races until \colorbox{highlightcolor}{the beginning of November}. \\
      \midrule
      \textbf{DAT$^{*}$ Vertex Predictions} & <BOS> There are Seven Seven There By seven races By are The are been seven races scheduled As until \textbf{the} \textbf{beginning} by \textbf{the} early early \textbf{beginning} \textbf{of} \textbf{November} early \textbf{November} \textbf{of} \textbf{November} , there will are are be have seven been up be are seven seven appear@@ seven races races have seven are races run are been scheduled p@@ place until play be scheduled take place until \textbf{the} \textbf{beginning} \textbf{beginning} \textbf{beginning} \textbf{of} in early \textbf{of} early early \textbf{November} . <EOS>  \\
      \textbf{DAT$^{*}$ Translation} &  There are seven races until \colorbox{highlightcolor}{the beginning of November}. \\ 
      \midrule
   \multicolumn{2}{c}{\textbf{Case 2}} \\
    \midrule
      \textbf{Source Sentence} & Bei der Kollision fliegen Hand- und Fußbremshebel weg.  \\
      \midrule
      \textbf{Reference Sentence} &  When they collided hand and foot brake pedals break off. \\
      \midrule
      \textbf{DAT Vertex Predictions} &  <BOS> 
      \textbf{During} The Hand Flying Hand@@ brake and In case \textbf{During} \textbf{the} case \textbf{During} \textbf{the} event of \textbf{the} \textbf{col@@ col@@} Col@@ ding \textbf{col@@ col@@ sion li@@ sion col@@ }\textbf{li@@ li@@ sion , sion} , \textbf{the} is \textbf{li@@} des leaves fly , the Hand le@@ es Hand@@ away Hand held of Hand of hand hand hand@@ wr@@ held hand and hand le@@ hand le@@ ver ver and ver foot le@@ ver ver and foot b@@ le brake foot foot brake foot bra@@ k@@ king brake brake brake le@@ vers fly le@@ le@@ le@@ le@@ vers vers vers are vers are fly fly fly fle@@ vers fly fly flying fly from the away f@@ away away away \textbf{during the} event \textbf{col@@ li@@ ding col@@ sion col@@ li@@ sion sion} <EOS> \\
      \midrule
      \textbf{DAT Translation} & \colorbox{highlightcolor}{During the collision}, hand and foot brake levers fly away \colorbox{highlightcolor}{during the collision}. \\
      \midrule
      \textbf{DAT$^{*}$ Vertex Predictions} & <BOS> Hand@@ -@@ Hand Hand@@ Flying In brake -@@ \textbf{During} and The foot \textbf{col@@} le@@ vers away Col@@ \textbf{during the} event of \textbf{the col@@ li@@} ding \textbf{col@@ col@@ col@@ sion li@@ sion li@@ sion} , \textbf{li@@} breaks involves session \textbf{li@@ li@@ sion} there , fly brake re@@ moves fly away of the Hand@@ Hand vers by Hand hand hand@@ held of hand hand hand le@@ - and hand brake brake hand le@@ and vers and and foot F@@ foot foot oot and foot foot under@@ king brake brake brake brake le@@ le@@ bra@@ vers arms vers are fly col@@ le@@ vers vers fly are fly fly flying f@@ fly ail away \textbf{during} away \textbf{the col@@} A@@ way away away \textbf{during the col@@ col@@ li@@ sion li@@} way <EOS>  \\
      \textbf{DAT$^{*}$ Translation} &  \colorbox{highlightcolor}{During the collision}, hand and foot brake levers fly away. \\   
    \bottomrule
    \end{tabular}
    \caption{
    \label{tab:dat_vertex_case}
    A case study of vertex predictions of DAT and DAT$^*$ models.
    The text in the grey background denotes the repeated segment in DAT.
    Tokens in bold denote the set of related vertex predictions that construct the phrase ``the begging of November''.
    generating repeated n-grams
    via finding a better vertex transition path, due to its stronger dependency modelling.
    }
    
\end{table*}

\section{Model Details of DAT$^*$}
\label{app:dat_star}
To strengthen the inter-token dependency of DAT, we increase the depth of the transition model by encoding $\mathbf{Q}$ in Equation~\ref{eq_qk} with an additional linear layer:
\begin{gather}
\label{eq:dat_refine}
    \mathbf{{Q}}^{*} = \text{ReLU}(\mathbf{{Q}}) \mathbf{W}_\text{Q}^{*},
\end{gather}
where ReLU is the rectified linear unit activation function.
The same applies to $\mathbf{K}$.
We refer to this model as DAT$^{*}$.

\section{Cross-domain Performance}
\label{app:cd}
The complete cross-domain performance on 5 De$\Rightarrow$En out-of-domain testsets are presented in Table~\ref{tab:ood}.

\begin{table*}[t]
\centering
\small
\begin{tabular}{lcccccc}
\toprule
\textbf{Model} & \multicolumn{1}{l}{\textbf{IT}} & \multicolumn{1}{l}{\textbf{Koran}} & \multicolumn{1}{l}{\textbf{Law}} & \multicolumn{1}{l}{\textbf{Medical}} & \multicolumn{1}{l}{\textbf{Subtitles}} & \multicolumn{1}{l}{\textbf{Average}}  \\
\midrule
       AT & \textbf{78.29} &  \textbf{62.08} & \textbf{85.44} &    \textbf{79.25} &      \textbf{75.09} & \textbf{76.03} \\
NAT & 32.97 &  33.10 & 34.53 &   32.61 &      34.90 & 33.62 \\
       MgMO & 72.39 &  57.89 & 79.00 &    73.27 &      69.24 & 70.36 \\
       CTC & 50.39 &  46.66 & 49.71 &    57.20 &      73.10 & 55.41 \\
         DAT & 77.86 &  61.15 & 85.07 &    78.78 &      74.76 & 75.52 \\
         CMLM & 61.67 &  38.30 & 71.94 &    59.97 &      47.42 & 55.86 \\
         DAT* & 77.57 &  61.33 & 84.05 &    78.78 &      74.65 &75.28 \\
         
\bottomrule
\end{tabular}
\caption{Cross-domian translation performance (COMET). Bold numbers represent the best performance.
\label{tab:ood}
}
\end{table*}
Compositional generalization in NMT refers to the model's generality to translate compounds (e.g., phrases) of known semantic units (e.g., words).
We test AT and DAT on the CoGnition dataset since they do not rely on knowledge distillation, and present the results in Table~\ref{tab:cg}.
As shown, DAT underperforms the AT counterpart in compositional generalization by a considerable margin, due to its weak dependency modeling.
DAT$^*$

\section{Robustness to Noisy Input}
\label{app:noise}
The translation performance on the WMT21 De$\Rightarrow$En testsets with different types of noises are shown in Table~\ref{tab:noise}, where ``None'' denotes the performance on the original testset without noise.

\begin{table*}[t]
    \centering
    \small
    \begin{tabular}{lcccc}
    \toprule
       \textbf{Model} & \textbf{None} & \textbf{Replace} & \textbf{Delete} & \textbf{Permutation} \\
    \midrule
        AT & 84.72 &  63.93 (-20.79) & 72.61 (-12.11) &  62.63 (-22.09)\\
        NAT & 75.50 &  57.38 (-18.12)& 64.36 (-11.14)& 53.28 (-22.22)\\
        CMLM  & 77.14 & 58.53 (-18.61)&  64.67 (-12.47)& 56.92 (-20.22)\\
        CTC & 80.06 & 64.27 (-15.79)&70.44 (-9.62) &60.65 (-19.41) \\
        MgMO & 81.15 & 64.86 (-16.29)& 71.09 (-10.06)& 61.15 (-20.00)\\
        DAT &  83.32 &64.02 (-19.30)& 71.28 (-12.04)& 60.79 (-22.53) \\
        DAT* & 83.20&64.75 (-18.45) &71.30 (-11.90) &60.59 (-22.61) \\
\bottomrule
\end{tabular}
\caption{Results of translation performance (COMET) on noisy testsets of WMT21 De$\Rightarrow$En.
\label{tab:noise}
}
\end{table*}

\end{document}